\documentclass{article}
\usepackage{graphicx}
\usepackage{multicol}
\usepackage{amsmath}

\usepackage[final]{neurips_2023}

\usepackage[utf8]{inputenc} 
\usepackage[T1]{fontenc}    
\usepackage{hyperref}       
\usepackage{url}            
\usepackage{booktabs}       
\usepackage{amsfonts}       
\usepackage{nicefrac}       
\usepackage{microtype}      
\usepackage{xcolor}         

\title{TreeFormers - An Exploration of Vision Transformers for Deforestation Driver Classification}

\author{%
  Uche Ochuba \\
  Department of Computer Science\\
  Stanford University\\
  Stanford, CA 94305 \\
  \texttt{uochuba@cs.stanford.edu} \\
}

\begin{document}

\maketitle

\begin{abstract}
  This paper addresses the critical issue of deforestation by exploring the application of vision transformers (ViTs) for classifying the drivers of deforestation using satellite imagery from Indonesian forests. Motivated by the urgency of this problem, I propose an approach that leverages ViTs and machine learning techniques. The input to my algorithm is a 332x332-pixel satellite image, and I employ a ViT architecture to predict the deforestation driver class; grassland shrubland, other, plantation, or smallholder agriculture. My methodology involves fine-tuning a pre-trained ViT on a dataset from the Stanford ML Group, and I experiment with rotational data augmentation techniques (among others) and embedding of longitudinal data to improve classification accuracy. I also tried training a ViT from scratch. Results indicate a significant improvement over baseline models, achieving a test accuracy of 72.9\%. I conduct a comprehensive analysis, including error patterns and metrics, to highlight the strengths and limitations of my approach. This research contributes to the ongoing efforts to address deforestation challenges through advanced computer vision techniques.
\end{abstract}

\section{Introduction}

This project targets the pressing issue of deforestation, which has a global negative impact on climate change, natural ecosystems, biodiversity, and habitat loss. Deforestation has been exacerbated by humans with many actions leading to forest area loss. Timely and accurate identification of the drivers of deforestation events is crucial (Austin et al., 2019). This project aims to explore the classification of deforestation drivers using satellite imagery from forests in Indonesia. The paper is motivated by the urgency of this problem and a strong desire based on past experiences to explore the potential benefits of employing vision transformers (ViT) and machine learning (ML) techniques.

The input to my algorithm is a 332x332-pixel satellite image of a site experiencing deforestation in Indonesia. I then use a ViT architecture to output a predicted class label for the deforestation driver, grassland shrubland, other, plantation, or smallholder agriculture.

\section{Related Work}

Typically, governments still use manual, expert annotation to classify and understand the drivers of deforestation. Naturally, this is slow and laborious, so there is a large opportunity for ML to improve this process. All current papers on this problem, to the best of my knowledge, use convolutional neural network architectures (CNNs) to address this problem. The first three papers discussed use the same dataset as this paper. This paper was initially inspired by \textit{Rotation Equivariant Deforestation Segmentation and Driver Classification} (Mitton \& Murray Smith, 2021). This paper is in the group of papers using CNNs to classify the images, including the use of CNNs with rotation equivariance, meaning they produce the same embedding for images irrespective of rotation. This is clever since we would not want a rotated image to cause the output of a different label (orientation is arbitrary for images of Earth). This paper however only reaches 63.0 accuracy on the test set. A second paper that uses CNNs is the source of the dataset, which uses an architecture called “ForestNet”, which combines a CNN with “scene data augmentation (SDA) where [they] randomly sample from the scenes and composite images during training to capture changes in the landscape over time”, and pre-training (PT) on other satellite images (Irvin et al., 2020). This paper is insightful as it uses composite imaging to capture change over time, a method that this paper also explores. Third, there is the paper Multimodal SuperCon: classifier for drivers of deforestation in Indonesia (Hartani et al., 2023) which builds upon the Mitton \& Murray paper by using a two-step training process called SuperCon (Chen et al., 2022), which first tries to learn features closely associated within one class and those that are more distant from other classes, and then fine tunes a classifier that makes predictions based on such a representation. This is an enlightening approach, as it has been shown to improve models on datasets with class imbalances, such as the dataset used in these papers. The data augmentation techniques in this paper incorporate elevation data, from the locations of the images. This paper also demonstrated an advancement in accuracy on the test set and was perhaps the most innovative approach that I studied.

I also looked at papers solving the same classification problem with CNNs on different datasets. Fourth, I looked at a paper by Pratiwi et al., 2021, which aims to use a four-layer CNN to classify images on a different, smaller dataset. They were able to achieve very respectable validation accuracy at on this dataset, which demonstrated the applicability of these methods to other datasets, as well as showing how ML models could be used to identify deforestation still in its early stages. Fifth, I looked at a paper by Subhahan \& Kumar, 2023, which was clever in that it used a forest of CNNs to provide classification results. This use of forests was interesting since it is to the best of my knowledge the only paper on this problem to use such a methodology, and as discussed in CS229 lecture weak learners can be combined to create a stronger learner, balancing the flaws of one another.

\section{Dataset and Features}

The dataset used is from the Stanford ML Group (Irvin et al., 2020), containing 2756 (1615 train (59\%), 473 validation (17\%), 668 test (24\%)) satellite images taken from 2001-16, all over the islands of Indonesia. For locations that have multiple images taken over time, the dataset has composites that blend the pixel intensity of all images over time. This captures the temporal nature of the data in a single image. The images are 332x332 pixels in a PNG format, and they are composite images of a given piece of land, taken over time. They are labeled by experts based on the driver of deforestation present in the image, either grassland shrubland, other, plantation, or smallholder agriculture. In terms of pre-processing, I wrote scripts to analyze the mean and standard deviation statistics of the dataset to apply normalization. I also created scripts to organize the data into an architecture based on the data splits and labels such that they could be passed into my data loader. The features used consist of RGB pixel intensities of the images.

\begin{figure}
  \centering
  \includegraphics[width=4cm, height=4cm]{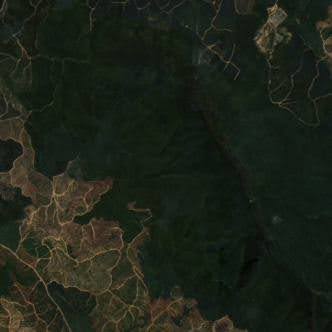}
  \includegraphics[width=4cm, height=4cm]{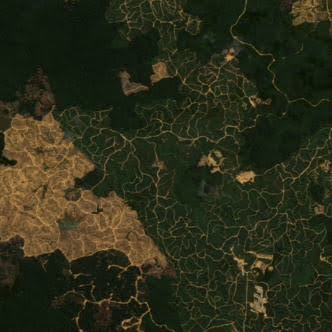}
  \includegraphics[width=4cm, height=4cm]{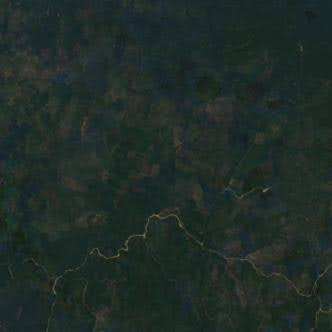}
    \includegraphics[width=4cm, height=4cm]{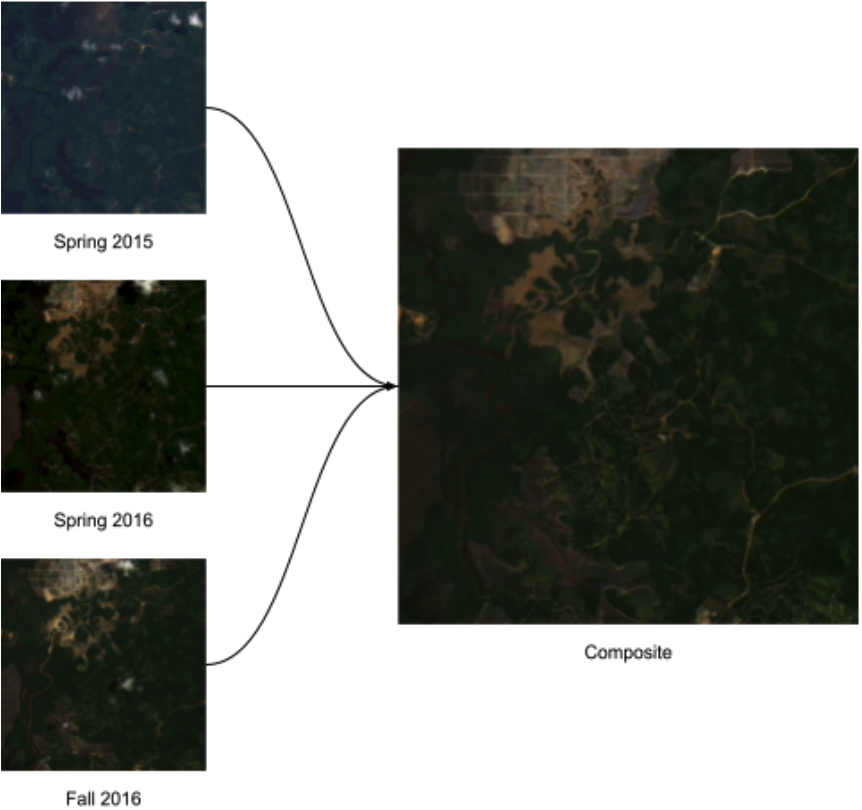}
    \caption{Some sample images from the dataset, a visualization of compositing of images.}
\end{figure}

Besides the regular vision transformer method, I implemented data augmentation techniques. Due to past promising results from apply rotation to this dataset (Mitton \& Murray-Smith, 2021), I added a model that has on-the-fly (OTF) data augmentation that applies a horizontal or vertical flip to the images, each with 50\% probability. I also created an “augmented model” which, in addition to the flipping behavior, applies the following data augmentations OTF during training: gray scaling with a 3
\begin{figure}
    \centering
    \includegraphics[width=3cm, height=1.5cm]{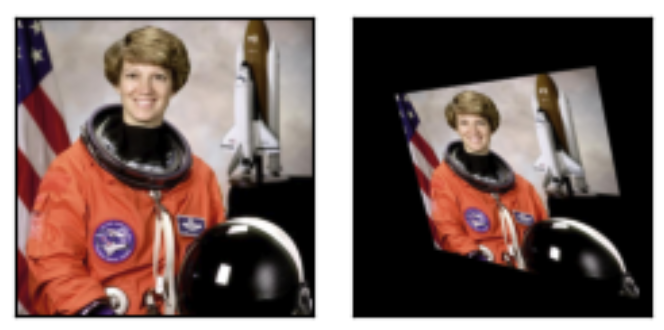}
    \includegraphics[width=3cm, height=1.5cm]{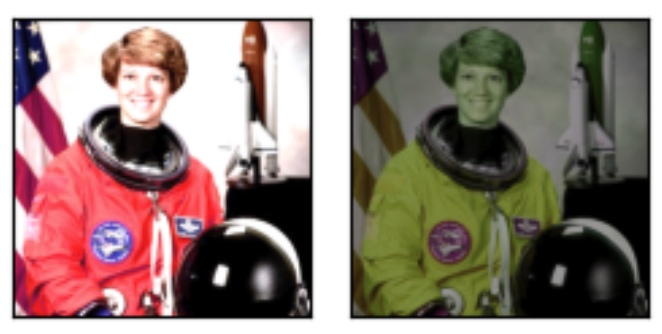}
    \includegraphics[width=1.5cm, height=1.5cm]{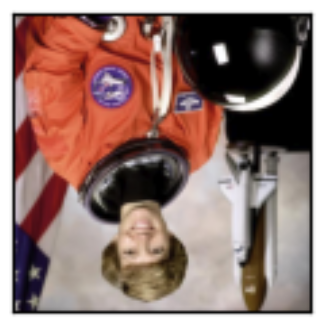}
    \caption{Visualizations of perspective change, color jitter, and flip/rotation transforms (Paszke et al., 2019).}
\end{figure}
In order to explore the data further, I constructed a t-distributed Stochastic Neighbor Embedding. This is a dimensionality reduction algorithm commonly used in vision papers to execute non-linear dimensionality reduction. The algorithm reduces the dimensionality of the images via a derivation obtained from KL-divergence, represented by $C$, which is obtained by the following equation: $C = D_{KL} (P||Q) = \sum_i \sum_j p_{ij} \log (\frac{p_{ij}}{q_{ij}})$, for two probability distributions, P and Q. In short, gradient updating is conducted with the following step for the images, to create a 2-dimensional embedding: $\mathcal{Y}^{(t)} = \mathcal{Y}^{(t-1)} + \eta \frac{\partial C}{\partial \mathcal{Y}} + \alpha (t) \left( \mathcal{Y}^{(t-1)} - \mathcal{Y}^{(t-2)}\right) $ where $\mathcal{Y}^{(t)}$ represents the solution at iteration $t$, $\eta$ represents the learning rate, and $\alpha (t)$ represents the momentum at iteration $t$ (Van der Maaten \& Hinton, 2008). Figure 3 shows the tSNE embedding for the raw pixel intensities dataset. It does not yield any significant insights, showing that this is a classification problem that is not easily separable by simpler ML methods.

\begin{figure}
    \centering
    \includegraphics[width=4.5cm, height=4.5cm]{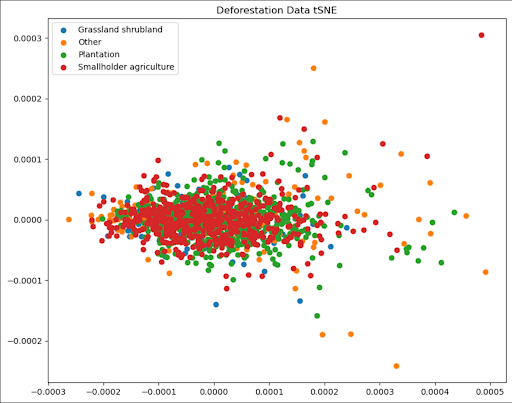}
    \includegraphics[width=5cm, height=4.5cm]{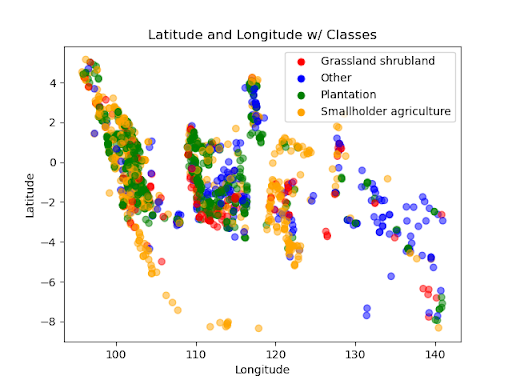}
    \includegraphics[width=4.5cm, height=4.5cm]{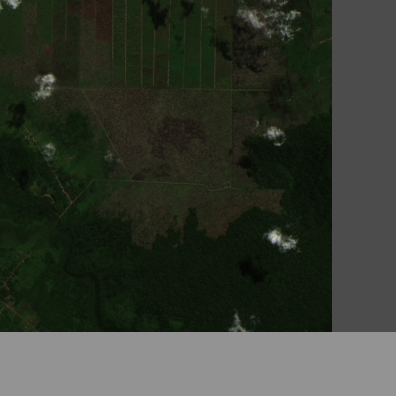}
    \includegraphics[width=4.5cm, height=4.5cm]{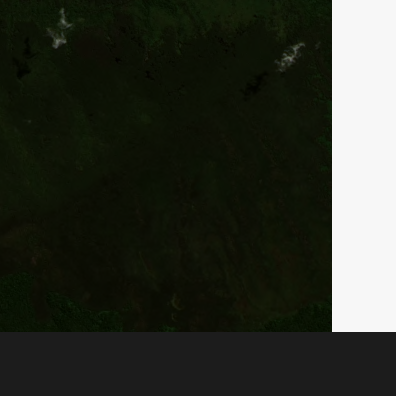}
    \caption{Left to right: tSNE visualization of image embeddings, plot of longitude vs. latitude with class labels, sample images with colored bars to embed longitudinal data.}
\end{figure}

After obtaining the dataset, I noticed that it naturally included latitudinal and longitudinal data, so I decided to explore if these features could be used to increase accuracy. In my exploration, I created a plot of the images’ latitude and longitude along with their class labels (see figure 3).

I noticed that there seemed to be some locations where specific labels were more clustered, so I decided to try to use longitudinal data as an input feature. As discussed further in the methods section, I tried methods that involved both using latitude and longitude as inputs in the classification head, as well as embedding the latitude and longitude as colored bars with different intensities based on the normalize latitude/longitude value. Such an image is shown in Figure 3. The satellite image remains in the top left, while the horizontal bar at the bottom’s intensity is chosen based on the value of the normalized longitude, and the vertical bar on the right’s intensity is brighter if there is a greater latitudinal value.
Finally, I wrote scripts to test each model on the test set, using the logits to create predictions, plots, confusion matrices, and probability distributions for each image, (probability $p$ calculated from the logits $L$ as follows: $p = \frac{1}{1+e^{-L}}$.

\section{Methods}

\subsection{Baselines}

To establish a baseline for testing, I implemented, from scratch, a logistic regression classifier for the images, using pixel intensities flattened into a vector. The logistic regression uses cross-entropy loss $\ell_\textup{CE}(\bar{h}_{\theta}(x),y) = - \log\left(\frac{\exp(\bar{h}_{\theta}(x)_{y})}{\sum_{s=1}^{k}\exp({\bar{h}_{\theta}(x)}_s)}\right)$, where h is the logits, y is the relevant coordinate of the vector, and the softmax function, $\sigma(z_i) = \frac{e^{z_{i}}}{\sum_{j=1}^K e^{z_{j}}}$ for a vector $z$, (commonly used to convert a vector of raw scores or logits into a probability distribution) for classification to perform four-class classification. Logistic regression is optimized using gradient descent. I also tried building my own "Custom ViT" without pre-training.

I also collected and ran the code used in the Mitton \& Murray–Smith paper to create a CNN baseline for the project, since so many past papers use this method. This CNN performs convolutions to reduce images to key features, and consists of layers with learnable filters or kernels that convolve over input data to extract local features. A U-Net architecture (Ronneberger et al., 2015) is used followed by ReLU ($ReLU(z) = \max(0, z)$) and pooling layers to downsample spatial dimensions by aggregating information. During training, the model adjusts its parameters using backpropagation updates.

\subsection{Vision Transformers}

This project mainly consists of an exploration into the application of vision transformers (ViTs) to this problem. Originally proposed by Dosovitskiy et al. in 2020, the ViT brings the power of transformers, widely used in natural language processing tasks, to the field of computer vision and image classification tasks. The ViT works by taking multiple fixed-sized, non-overlapping patches of the image to later create attentionality, where the model is able to “focus” on different parts of the image. The algorithm passes a linear projection of the patches into the transformer encoder (the projection of a vector $v$ onto a vector $y$ is $\frac{v \cdot y}{v \cdot y} y$. LayerNorm is applied to these data $y= \frac{x - E[x]}{\sqrt{Var[x] + \epsilon}} \cdot \gamma \beta$ (where $y$ is the output tensor, $x$ is the input tensor, $E[x]$ is the expectation of $x$, $Var[x]$ is the variance of $x$ and $\gamma, \beta$ are learned transform parameters), followed by a multi-head self-attention layer, and another LayerNorm layer. The last part of the transformer encoder is a multilayer perception (MLP) layer, which consists of individual processing units, termed "neurons," arranged hierarchically. Mathematically, this involves multiplying each vector $v$ of the layer by a normalized version of $\text{softmax} (qk^T)$, where $q, k$ are vectors "adjacent" to $v$. Each neuron in a layer receives input from all neurons in the preceding layer through weighted connections. The layer applies a transformation to the input data using the GeLU function, $GeLU(x) = x\Phi(x)$, (where $\Phi(x)$ is the cumulative distribution function of the Gaussian distribution) with their corresponding weights and summing them up, allowing the network to progressively extract higher-level representations as it propagates through the layers (Dosovitskiy et al., 2020). Finally, more GeLU functions are applied to the output of the encoder with the MLP head, and along with the softmax function, the logits are produced for the classification task output.

Vision transformers are often pre-trained on extremely large datasets, and then fine-tuned for specific classification tasks. In this project, I tried building and training a vision transformer from scratch based on code from a paper by Li et al., 2021, calling this the custom ViT. Then, for the bulk of the exploration, I eventually pursued the path of fine-tuning a pre-trained transformer (Steiner et al., 2021). This was trained on the ImageNet-21k dataset, which contains 14,197,122 images divided into 21,841 classes (Ridnik et al., 2021). I iterated upon the fine-tuning process for these models, experimenting with different batch sizes, including adding rotation/flips and data augmentation.

Next, I wanted to find a methodology to incorporate the longitudinal data into the training and prediction process. To do so, I proposed a PyTorch architecture which would take the information from the final hidden layer of the transformer encoder, and then pass in longitudinal data before the execution of the MLP head with GeLUs. The proposed architecture is shown in Figure 4, although this approach was ultimately unsuccessful.

\begin{figure}
    \centering
    \includegraphics[width=6cm, height=3cm]{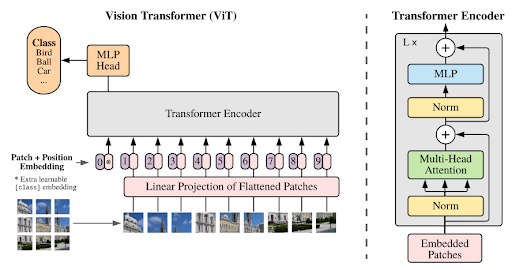}
    \includegraphics[width=4.5cm, height=3cm]{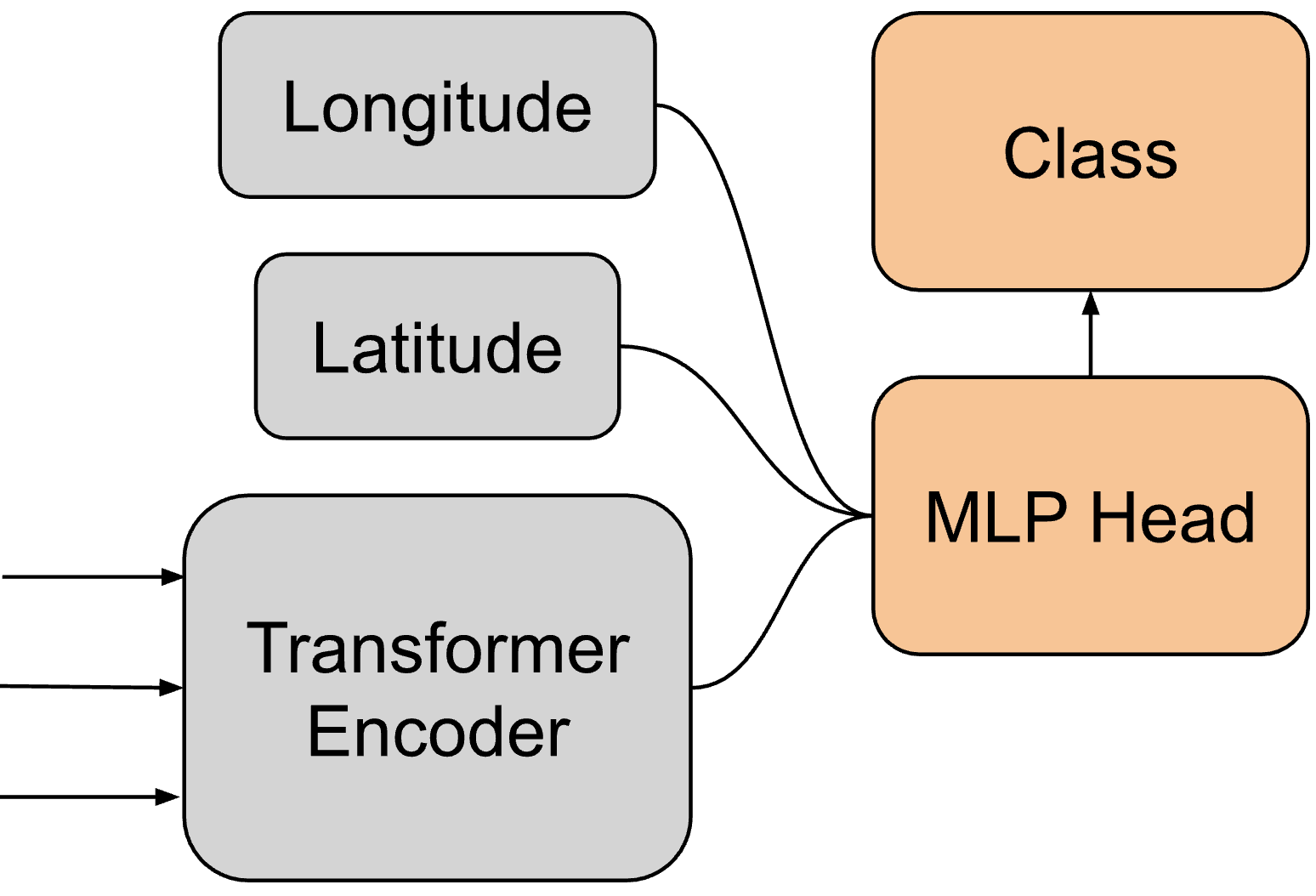}
    \caption{Left to right: The ViT architecture (Dosovitskiy et al., 2020), the proposed modified classification head architecture to incorporate longitudinal data into predictions.}
    \label{fig:enter-label}
\end{figure}

Given this, I looked to explore new methods to embed the longitudinal data into the transformer. I decided to use bars with varying intensities based on the normalized latitude and longitude of the data. This led to images as shown in the dataset and features section, with vertical and horizontal colored bars. I also experimented with using such images with data augmentation techniques. 

\section{Experiments/Results/Discussion}

\begin{table}
  \caption{Model Performance Comparison}
  \label{model-performance}
  \centering
  \begin{tabular}{lcc}
    \toprule
    Model & Top 1 Acc (Val) & Top 1 Acc (Test) \\
    \midrule
    Naive LR Baseline & 35.0 & 33.3 \\
    UNET - CNN & 60.6 & 57.9 \\
    UNET - C8 Equivariant & 67.1 & 63.0 \\
    Custom ViT & 45.4 & 38.4 \\
    ViT & 79.1 & 72.0 \\
    \textbf{Rotation ViT} & \textbf{80.1} & \textbf{72.9} \\
    Augmented ViT & 79.5 & 72.9 \\
    Longitudinal ViT & 77.6 & 71.7 \\
    Longitudinal+Rotation ViT & 79.5 & 72.3 \\
    Longitudinal+Augmented ViT & 79.3 & 72.6 \\
    \bottomrule
  \end{tabular}
\end{table}

\begin{table}
  \caption{Model Performance Metrics}
  \label{model-metrics}
  \centering
  \begin{tabular}{lccccc}
    \toprule
    Model & Precision & Recall & F1 & AUROC & AUPRC \\
    \midrule
    ViT & 0.694 & 0.649 & 0.664 & 0.868 & \textbf{0.727} \\
    \textbf{Rotation ViT} & 0.703 & \textbf{0.661} & \textbf{0.676} & \textbf{0.877} & 0.718 \\
    Augmented ViT & \textbf{0.711} & 0.654 & 0.671 & 0.875 & 0.718 \\
    Longitudinal ViT & 0.687 & 0.649 & 0.660 & 0.852 & 0.692 \\
    Longitudinal+Rotation ViT & 0.695 & 0.655 & 0.666 & 0.865 & 0.692 \\
    Longitudinal+Augmented ViT & 0.716 & 0.652 & 0.670 & 0.867 & 0.714 \\
    \bottomrule
  \end{tabular}
\end{table}

\begin{figure}
    \centering
    \includegraphics[width=6cm, height=3cm]{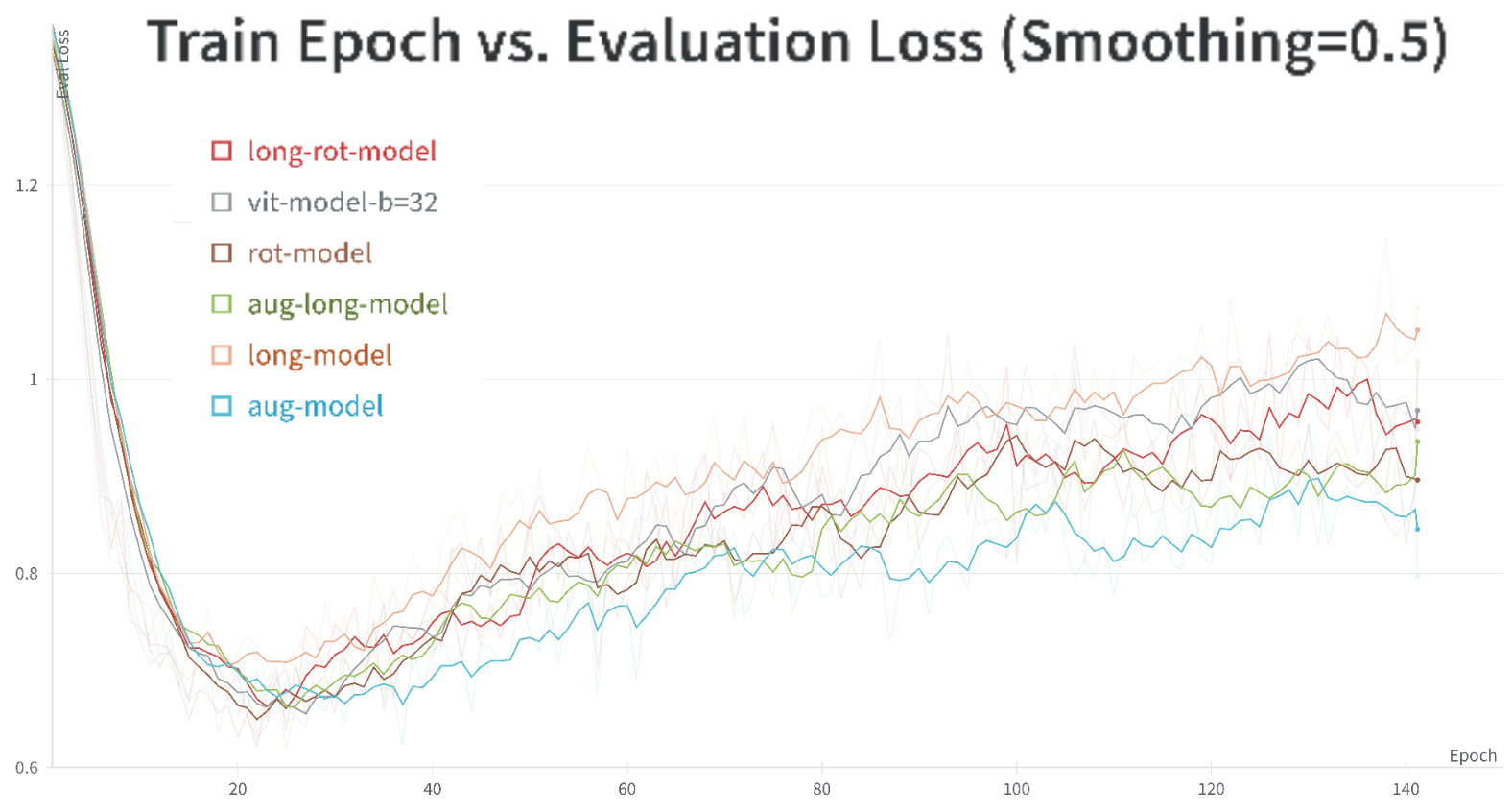}
    \includegraphics[width=6cm, height=3cm]{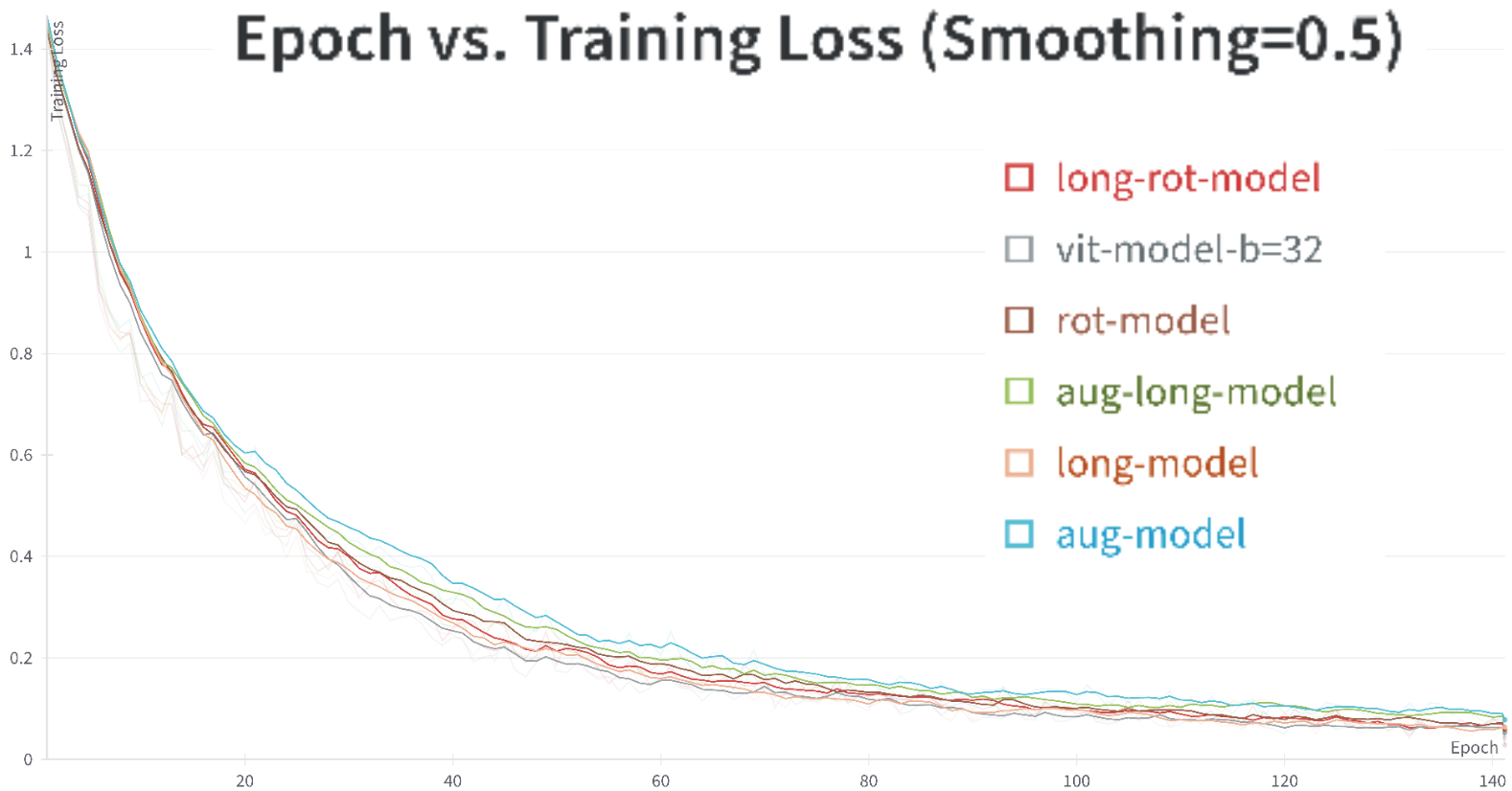}
    \includegraphics[width=6cm, height=3cm]{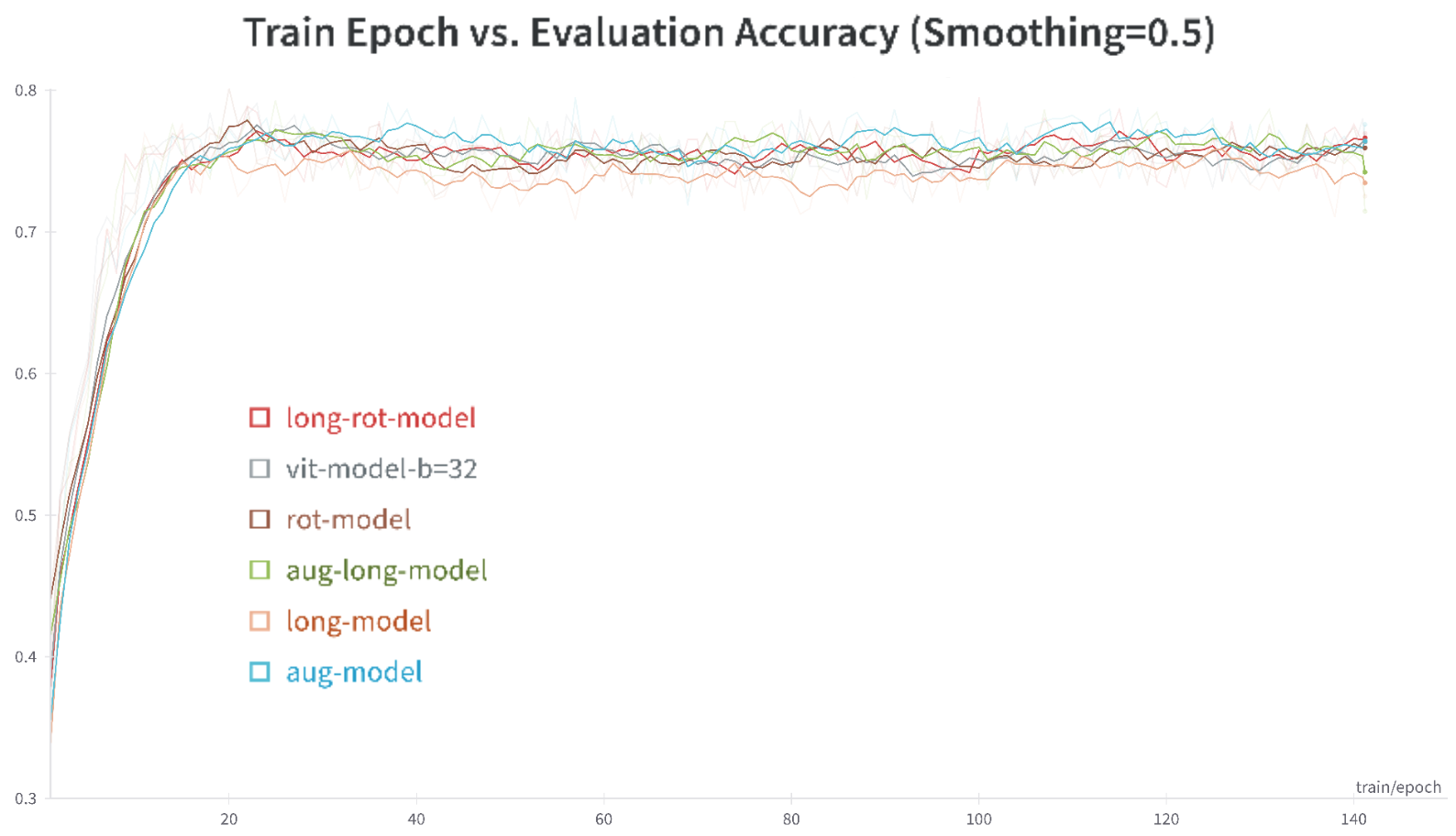}
    \caption{Statistics over 300 eopchs for various models. The highest y-tick-marks for figures are 1.2, 1.4, and 0.8, from left to right.}
\end{figure} 

Models were fine-tuned with NVIDIA T4 GPUs for 150 epochs, taking about 3 hours to train on average. I chose 150 epochs as I noticed that eval accuracy had well converged by this point. In preliminary investigations, I discovered that a batch size of 32 led to the best evaluation accuracy and smallest evaluation loss, while not running out of memory on the machines. Next, I chose a learning rate of $5 \times e^{-5}$. This was also chosen with experimentation and knowledge of the effects of different learning rates. I wanted to select a learning rate large enough such that the model would converge quickly enough, and one precise enough so as not to bounce around too erratically on the cost function surface. $5 \times e^{-5}$ provided the best of both worlds. The model I selected at the end of fine-tuning was that with the lowest evaluation loss. I chose this as I wanted to deal with the inherent class imbalance of the dataset, and I wanted to ensure that there was less bias toward simply making more classifications toward classes that had more examples in the overall dataset. Intuitively, these models also had amongst the highest evaluation accuracy of those in any epoch.

My primary metrics are standard for classification problems, including accuracy $=\frac{TP +TN}{TP + TN + TP + TN}$, precision $=\frac{TP}{TP+FP}$, recall $=\frac{TP}{TP+FN}$, F1 $=\frac{2 \cdot precision \cdot recall}{precision + recall}$, AUROC (the area under the receiver operating characteristic curve, as discussed in lecture), and AUPRC (the area under the precision-recall curve, as discussed in lecture). The last two are particularly useful for handling class imbalance in the dataset. Where we have the following counts: TP for true positives, TN for true negatives, FP for false positives, FN for false negatives. For multiclass properties, I calculated AUROC and AUPRC using one-vs-rest methodology.

The pre-trained ViT saw a test accuracy of 72.0\%, which increased to 72.9\% when adding in rotation-based data augmentations. The auxiliary data augmentations, which make up the “augmented model” in addition to rotations, decreased performance on the evaluation set and did not provide a meaningful increase in accuracy on the test set. This test set accuracy represents a vast improvement over the naive logistic regression baseline, as well as 9.9\% increase in accuracy compared to the results achieved by running the rotation equivariant CNN (Mitton \& Murray–Smith, 2021). This is also a 1.9\% accuracy to that achieved in the aforementioned 2023 SuperCon paper (Hartani et al., 2023). This approaches the 80.0\% accuracy of Irvin’s ForestNet, even without the auxiliary data that ForesNet uses (Irvin et al., 2021).

The logistic regression baseline saw a test accuracy of 33.3\%, which is close to the expectation of random guessing, which sits at 25.0\%. The custom, non-pre-trained ViT which was only (pre-)trained on forest satellite images, achieved 38.4\% accuracy. The inclusion of the longitudinal data in bar form seemed to hurt the accuracy of the predictions, and the custom MLP classifier with longitudinal data also presented deprecated accuracies, and so was not included. This lack of success with longitudinal data perhaps indicates that there was more noise in this signal than originally thought, or that the method of embedding the data into each image was flawed.

The success of the rotational ViT model is promising, attributable to the attention capabilities of vision transformers, as well as robust pre-training. The results also show, in line with previous approaches, that training methodologies that account for rotation to negate the arbitrary nature of the orientation of satellite images are successful for this classification problem (Mitton \& Murray–Smith, 2021). The rotation data augmentation mainly achieved its success by more accurately classifying images in the smallest cardinality class, grassland shrubland, without giving up accuracy in other categories. This model also shows promising statistics across F1, precision, recall, AUROC, and AUPRC, particularly across the plantation class.

\begin{figure}
    \centering
    \includegraphics[width=4cm, height=4cm]{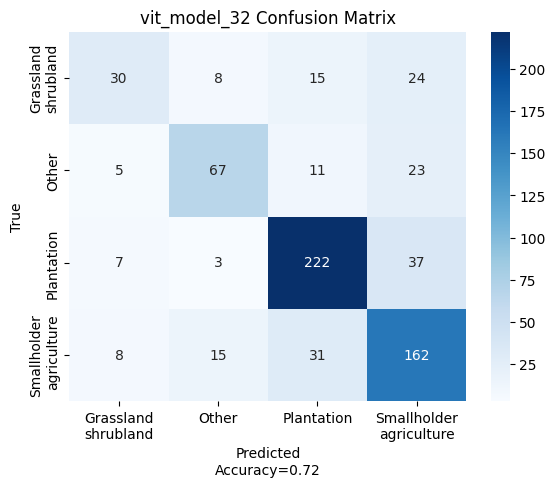}
    \includegraphics[width=4cm, height=4cm]{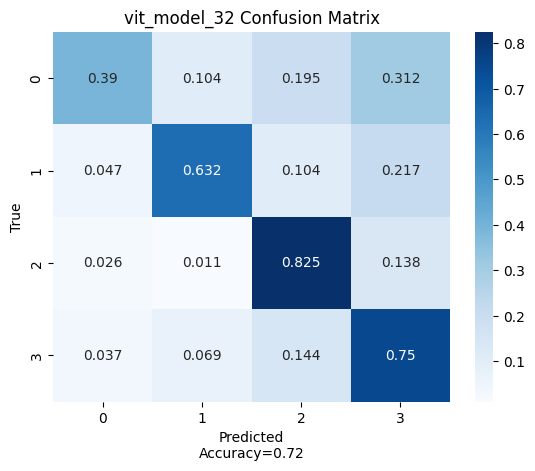}
    \includegraphics[width=4cm, height=4cm]{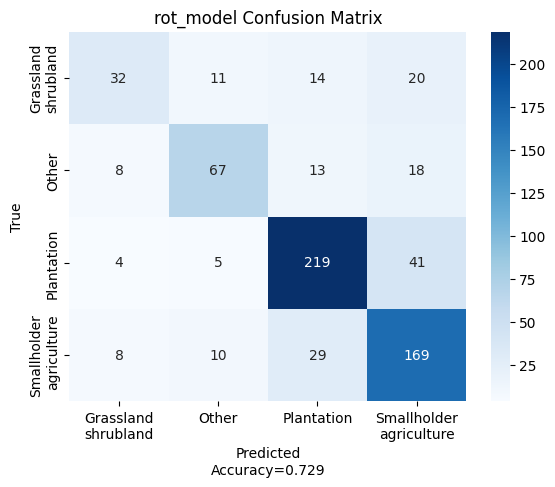}
    \includegraphics[width=4cm, height=4cm]{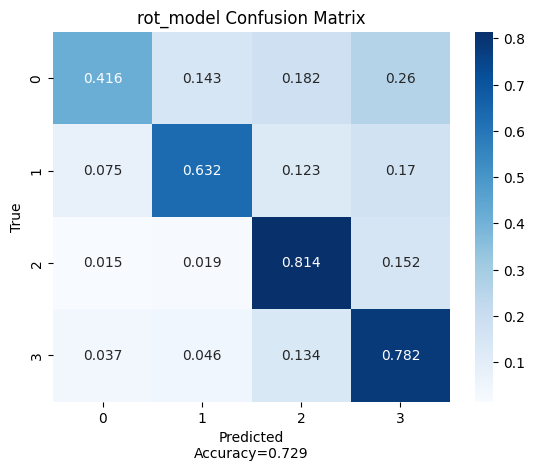}
    \caption{Left to right: confusion matrices for the regular ViT and the rotation/flip data-augmented ViT, with values absolute and normalized by true class label.}
\end{figure}

\begin{table}
  \caption{Model Performance Metrics for Rotation ViT}
  \label{model-metrics-rotation-vit}
  \centering
  \begin{tabular}{lccccc}
    \toprule
    Class & Precision & Recall & F1 & AUROC & AUPRC \\
    \midrule
    Grassland shrubland & 0.615 & 0.416 & 0.496 & 0.812 & 0.515 \\
    Other & 0.720 & 0.632 & 0.673 & 0.909 & 0.737 \\
    Plantation & 0.796 & 0.814 & 0.805 & 0.907 & 0.860 \\
    Smallholder agriculture & 0.681 & 0.782 & 0.728 & 0.879 & 0.761 \\
    \textbf{Mean} & \textbf{0.703} & \textbf{0.661} & \textbf{0.676} & \textbf{0.877} & \textbf{0.718} \\
    \bottomrule
  \end{tabular}
\end{table}

In terms of error analysis, we see that the rotation model still struggles to classify images that are in the grassland shrubland class, for which the dataset has the fewest images. Many such images were erroneously classified as smallholder agriculture by all vision transformers. This is an acute challenge that is noted in previous papers as well (Hartani et al., 2023). There seems to be a general trend toward the over-predicting of smallholder agriculture across all classes. I infer that this is because it is one of the largest cardinality sets in the training set, but also because the very largest cardinality set (plantation), has features that are much less similar to any of the other classes. As one might imagine, large-scale plantations tended to have features that were more recognizable to the vision transformers, as they tended to take up more space in any given image.

All of the pre-trained models saw training loss that decreased more or less monotonically, evaluation loss that bottomed out at around 20 epochs, then increased, and evaluation accuracy that peaked at around 20 epochs and then stayed steady. I do not think that there is overfitting to the training sets, as evaluation accuracy holds steady across all models even as I tune for a greater and greater number of epochs. I also chose to use the model with the lowest evaluation accuracy as the final model. Finally, data augmentation naturally helps to mitigate overfitting, as I am technically training on a newly modified dataset at each epoch.

\begin{figure}
    \centering
    \includegraphics[height=3cm, width=3cm]{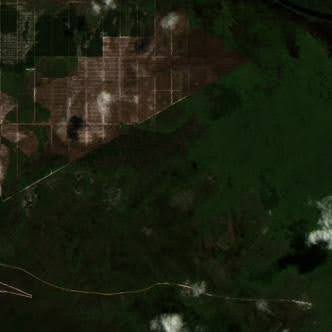}
    \includegraphics[height=3cm, width=3cm]{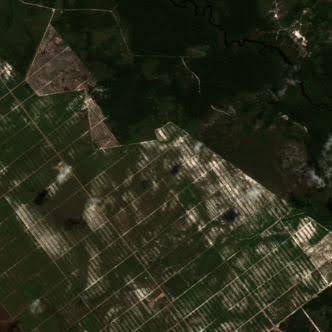}
    \includegraphics[height=3cm, width=3cm]{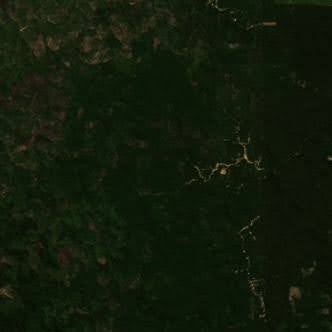}
    \caption{Left to right: a grassland shrubland image misclassified as smallholder agriculture. An image with label true 'plantation'. An image with true label 'smallholder agriculture'. While the latter two mentioned labels hold outsized cardinality in the dataset, it seems plantations are easier for the model to distinguish than smallholder agriculture.}
\end{figure}

\section{Conclusion/Future Work}

In conclusion, ViTs show much promise for working on deforestation image classification. I saw a 72.9\% accuracy for a pre-trained ViT aided by rotational data augmentation, an improvement on various past papers, especially those that only used image data, such as this one. This can be attributed to the power of multi-layered encoding in ViTs, as well as rotational data augmentations. This was better than the ViTs which incorporate longitudinal data, meaning that such information likely has less predictive power than originally thought.

In the future, I would want to further iterate on creating a custom MLP classification head to incorporate longitudinal data. I would also explore using different pre-trained vision transformers, or potentially curating topological or demographic data to incorporate into predictions to improve accuracy. I might also want to look into artificially re-balancing the dataset, or changing the data split to increase the size of the training set.

\section{Contributions}

This was a solo project, and thus all contributions were my own. Including various dataloaders, training and testing scripts, baseline logistics regression code, experimentation, and visualizations.

\section{References/Bibliography \& Acknowledgments}

I would like to thank my esteemed project advisor, Jeff Z. HaoChen, as well as my mom, dad, and brothers for their support. \\

[1] Abdus Subhahan, D., \& Vinoth Kumar, C. N. S. (2023). Classifying drivers of deforestation by using the deep learning based poly-highway forest convolution network. \emph{Journal of Intelligent \& Fuzzy Systems}, Preprint(Preprint), 1–15. \url{https://doi.org/10.3233/JIFS-233534}

[2] Austin, K. G., Schwantes, A., Gu, Y., \& Kasibhatla, P. S. (2019). What causes deforestation in Indonesia? \emph{Environmental Research Letters}, 14(2), 024007. \url{https://doi.org/10.1088/1748-9326/aaf6db}

[3] Biewald, L. (2020). Experiment Tracking with Weights and Biases. \url{https://www.wandb.com/}

[4] Chen, K., Zhuang, D., \& Chang, J. M. (2022). SuperCon: Supervised Contrastive Learning for Imbalanced Skin Lesion Classification (\emph{arXiv:2202.05685}). \emph{arXiv}. \url{https://doi.org/10.48550/arXiv.2202.05685}

[5] Clark, J. A. (n.d.). Pillow (PIL Fork) Documentation.

[6] Dosovitskiy, A., Beyer, L., Kolesnikov, A., Weissenborn, D., Zhai, X., Unterthiner, T., Dehghani, M., Minderer, M., Heigold, G., Gelly, S., Uszkoreit, J., \& Houlsby, N. (2021). An Image is Worth 16x16 Words: Transformers for Image Recognition at Scale (\emph{arXiv:2010.11929}). \emph{arXiv}. \url{https://doi.org/10.48550/arXiv.2010.11929}

[7] Harris, C. R., Millman, K. J., van der Walt, S. J., Gommers, R., Virtanen, P., Cournapeau, D., Wieser, E., Taylor, J., Berg, S., Smith, N. J., Kern, R., Picus, M., Hoyer, S., van Kerkwijk, M. H., Brett, M., Haldane, A., del Río, J. F., Wiebe, M., Peterson, P., … Oliphant, T. E. (2020). Array programming with NumPy. \emph{Nature}, 585(7825), Article 7825. \url{https://doi.org/10.1038/s41586-020-2649-2}

[8] Hartanti, B. S. I., Vito, V., Arymurthy, A. M., Krisnadhi, A. A., \& Setiyoko, A. (2023). Multimodal SuperCon: Classifier for drivers of deforestation in Indonesia. \emph{Journal of Applied Remote Sensing}, 17(3), 036502. \url{https://doi.org/10.1117/1.JRS.17.036502}

[9] Hunter, J. D. (2007). Matplotlib: A 2D Graphics Environment. \emph{Computing in Science \& Engineering}, 9(3), 90–95. \url{https://doi.org/10.1109/MCSE.2007.55}

[10] Irvin, J., Sheng, H., Ramachandran, N., Johnson-Yu, S., Zhou, S., Story, K., Rustowicz, R., Elsworth, C., Austin, K., \& Ng, A. Y. (2020). ForestNet: Classifying Drivers of Deforestation in Indonesia using Deep Learning on Satellite Imagery (\emph{arXiv:2011.05479}). \emph{arXiv}. \url{https://doi.org/10.48550/arXiv.2011.05479}

[11] Lhoest, Q., del Moral, A. V., Jernite, Y., Thakur, A., von Platen, P., Patil, S., Chaumond, J., Drame, M., Plu, J., Tunstall, L., Davison, J., Šaško, M., Chhablani, G., Malik, B., Brandeis, S., Scao, T. L., Sanh, V., Xu, C., Patry, N., … Wolf, T. (2021). Datasets: A Community Library for Natural Language Processing (\emph{arXiv:2109.02846}). \emph{arXiv}. \url{https://doi.org/10.48550/arXiv.2109.02846}

[12] Li, Y., Wu, C.-Y., Fan, H., Mangalam, K., Xiong, B., Malik, J., \& Feichtenhofer, C. (2022). MViTv2: Improved Multiscale Vision Transformers for Classification and Detection (\emph{arXiv:2112.01526}). \emph{arXiv}. \url{https://doi.org/10.48550/arXiv.2112.01526}

[13] Mitton, J., \& Murray-Smith, R. (2021). Rotation Equivariant Deforestation Segmentation and Driver Classification (\emph{arXiv:2110.13097; Version 2}). \emph{arXiv}. \url{http://arxiv.org/abs/2110.13097}

[14] Paszke, A., Gross, S., Massa, F., Lerer, A., Bradbury, J., Chanan, G., Killeen, T., Lin, Z., Gimelshein, N., Antiga, L., Desmaison, A., Köpf, A., Yang, E., DeVito, Z., Raison, M., Tejani, A., Chilamkurthy, S., Steiner, B., Fang, L., … Chintala, S. (2019). PyTorch: An Imperative Style, High-Performance Deep Learning Library (\emph{arXiv:1912.01703}). \emph{arXiv}. \url{https://doi.org/10.48550/arXiv.1912.01703}

[15] Pedregosa, F., Varoquaux, G., Gramfort, A., Michel, V., Thirion, B., Grisel, O., Blondel, M., Müller, A., Nothman, J., Louppe, G., Prettenhofer, P., Weiss, R., Dubourg, V., Vanderplas, J., Passos, A., Cournapeau, D., Brucher, M., Perrot, M., \& Duchesnay, É. (2018). Scikit-learn: Machine Learning in Python (\emph{arXiv:1201.0490}). \emph{arXiv}. \url{https://doi.org/10.48550/arXiv.1201.0490}

[16] Pratiwi, N. K. C., Fu’adah, Y. N., \& Edwar, E. (2021). Early Detection of Deforestation through Satellite Land Geospatial Images based on CNN Architecture. \emph{JURNAL INFOTEL}, 13(2), Article 2. \url{https://doi.org/10.20895/infotel.v13i2.642}

[17] Ridnik, T., Ben-Baruch, E., Noy, A., \& Zelnik-Manor, L. (2021). ImageNet-21K Pretraining for the Masses (\emph{arXiv:2104.10972}). \emph{arXiv}. \url{https://doi.org/10.48550/arXiv.2104.10972}

[18] Ronneberger, O., Fischer, P., \& Brox, T. (2015). U-Net: Convolutional Networks for Biomedical Image Segmentation (\emph{arXiv:1505.04597}). \emph{arXiv}. \url{https://doi.org/10.48550/arXiv.1505.04597}

[19] Steiner, A., Kolesnikov, A., Zhai, X., Wightman, R., Uszkoreit, J., \& Beyer, L. (2022). How to train your ViT? Data, Augmentation, and Regularization in Vision Transformers (\emph{arXiv:2106.10270}). \emph{arXiv}. \url{https://doi.org/10.48550/arXiv.2106.10270}

[20] team, T. pandas development. (2023). pandas-dev/pandas: Pandas (v2.1.4) [Computer software]. Zenodo. \url{https://doi.org/10.5281/zenodo.10304236}

[21] van der Maaten, L. J. P., \& Hinton, G. E. (2008). Visualizing High-Dimensional Data Using t-SNE. \emph{Journal of Machine Learning Research}, 9(nov), 2579–2605.

[22] Virtanen, P., Gommers, R., Oliphant, T. E., Haberland, M., Reddy, T., Cournapeau, D., Burovski, E., Peterson, P., Weckesser, W., Bright, J., van der Walt, S. J., Brett, M., Wilson, J., Millman, K. J., Mayorov, N., Nelson, A. R. J., Jones, E., Kern, R., Larson, E., … van Mulbregt, P. (2020). SciPy 1.0: Fundamental algorithms for scientific computing in Python. \emph{Nature Methods}, 17(3), Article 3. \url{https://doi.org/10.1038/s41592-019-0686-2}

[23] Waskom, M. L. (2021). seaborn: Statistical data visualization. \emph{Journal of Open Source Software}, 6(60), 3021. \url{https://doi.org/10.21105/joss.03021}

[24] Wolf, T., Debut, L., Sanh, V., Chaumond, J., Delangue, C., Moi, A., Cistac, P., Rault, T., Louf, R., Funtowicz, M., Davison, J., Shleifer, S., von Platen, P., Ma, C., Jernite, Y., Plu, J., Xu, C., Scao, T. L., Gugger, S., … Rush, A. M. (2020). HuggingFace’s Transformers: State-of-the-art Natural Language Processing (\emph{arXiv:1910.03771}). \emph{arXiv}. \url{https://doi.org/10.48550/arXiv.1910.03771}


\end{document}